\documentclass[letterpaper, 10 pt, journal]{ieeeconf}
\IEEEoverridecommandlockouts 
\usepackage{decar-common}
\usepackage{decar-dynamics}
\usepackage{decar-lie}
\usepackage{decar-post} 
\usepackage{graphicx}
\usepackage{lipsum}  
\usepackage{diagbox} 
\usepackage{booktabs} 
\usepackage{tabularx} 
\usepackage{makecell}

\graphicspath{{figs/}}

\title{\LARGE \bf
    D-CLIPSE: Distributed Consensus-based Localization with Passive Listening on Shared State Exchange
}

\author{Kyle Biron-Gricken and James Richard Forbes}

\begin{document}
    \maketitle
    \thispagestyle{empty}
    \pagestyle{empty}
    \begin{abstract}
    Multi-robot localization that is accurate and consistent is imperative for downstream tasks such as planning and control. Centralized filtering approaches optimally fuse all available sensor measurements of the team. However, a centralized solution is rarely implementable due to hardware, communication, and computational constraints. Distributed approaches deploy a filter on each robot to estimate their own state and neighbours' states using inter-robot communication. This paper proposes a consistent, communication-efficient, and consensus-based distributed filtering framework that shares both preintegrated odometry and relevant shared states among communicating robots. The proposed method is validated in simulated and experimental scenarios, showing near centralized performance in accuracy, and especially in consistency, compared to the current state-of-the-art decentralized approach. 
\end{abstract}    
    \section{Introduction}
Multi-robot applications are becoming increasingly prevalent in many industries, such as infrastructure inspection, forestry, and mining~\cite{foster_infrastructure_2024,
chen_forestry_2025,
vigara_puche_underground_2026}. To accomplish tasks successfully, a multi-robot team must have an accurate and consistent localization solution distributed among the team to inform downstream decisions, such as path planning and formation control. 

\begin{figure}[htbp]
    \centering
    \includegraphics[width=\linewidth, page = 6]{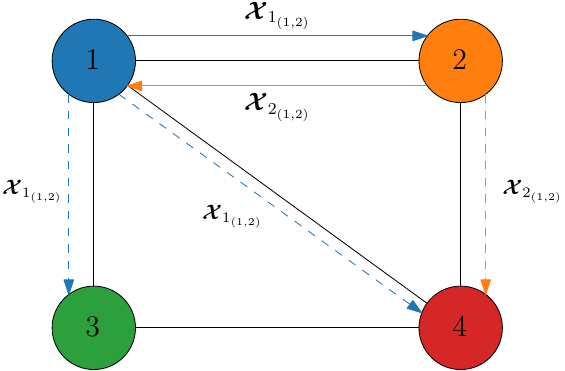}
    \caption{An example multi-robot communication graph, where Robot 1 and Robot 2 exchange shared states. Robots 3 and 4 passively listen in on the shared state exchange, and update their local estimates based on the overheard information.}
    \label{fig:comm_graph}
    \vspace{-3pt}
\end{figure}

One challenge of multi-robot navigation is that robots may be equipped with different sensing modalities. In some situations, a robot may not have enough sensor information to estimate its own local state, and therefore must rely on its neighbours for additional information. A \emph{centralized} approach to solving this problem aggregates all robots' local sensor data to a central processor, producing a state estimate of each member of the team with the lowest estimation error uncertainty. However, this approach is often unattainable in practice due to single-point-of-failure concerns, infrastructure limitations, and communication constraints. 
Instead, a \emph{distributed} solution that is local to each robot is desirable, as it enables the use of local storage and computation, and introduces redundancy. In a distributed setting, each robot estimates its own state or its own state in addition to the state of each of its neighbours. A benefit of estimating the neighbours' states is to inform local path planning and control decisions, while also locally tracking inter-robot cross-correlations. As shown in Fig.~\ref{fig:comm_graph}, this is realized through inter-robot communication, where robots share both their local sensor data and state estimates over some communication medium, such as Wi-Fi or ultra-wideband (UWB) radio. Additionally, \emph{passive listening} allows non-communicating neighbours to overhear shared information. Robots that estimate the same states motivates \emph{consensus}, such that the robots agree on the subset of the states that they share. By enforcing consensus between communicating robots, each robot's local estimate should converge toward the centralized solution, as they are updated with information from other robots' local estimates and sensor data. 
\subsection{Related Work}
Consensus-based filtering techniques have been studied extensively for problems pertaining to sensor networks and target-tracking~\cite{olfati-saber-distributed-2005,olfati-saber-distributed-2009, battistelli_consensus-based_2015}. In these scenarios, the estimator networks are static sensors that come to a consensus on the state of the dynamic target through local observations and information exchange. These methods consider sensors either sharing state estimates, measurement innovations, or a hybrid of both, over multiple, all-to-all communication rounds.  The above methods assume that the dynamics of the target and its associated process input and noise covariance are known to all sensors, and consensus is achieved through multiple all-to-all communication rounds. 

In terms of pairwise communication,~\cite{luft_recursive_2018} proposes robots locally track inter-robot cross-correlations to neighbouring poses. The work~\cite{jung_decentralized_2020} further applies this to a 3D visual-inertial-odometry (VIO) setting, in which all robots' poses are resolved in one common reference frame. The estimators in these methods only estimate and communicate their local pose estimates and neighbour cross-correlations, and do not have access to input information from other robots.

Other methods that do not explicitly track inter-estimator cross-correlations tend to rely on the provably consistent fusion method of \emph{covariance intersection} (CI)~\cite{julier_non-divergent_1997} when cross-correlations are unknown. The work~\cite{hao_decentralized_2025} proposes a two-step fusion method that requires, at minimum, two communication rounds with all interacting neighbours, where both fusion steps employ CI. The work~\cite{zhou_distributed_2026} extends upon this by considering a matrix Lie group (MLG) state definition based on VIO, where robots hold their local estimate, sharing it when relative measurements occur. In these methods, measurement information and local state estimates are shared, often with multiple communication rounds.

The proposed method presented in this paper is motivated by the work of~\cite{shalaby_multi_robot_2024} and~\cite{cossette_decentralize_2024}. The former considers multi-robot relative pose estimation through sharing preintegrated odometry, or \emph{relative motion increments} (RMIs), along with a passive listening-based framework for non-interacting neighbours to overhear shared RMIs. The latter is the closest to this paper's proposed method and considers a general decentralized filtering framework where through user-defined pseudomeasurements, sharing RMIs, and state sharing, multiple robots estimate their own state and the state of each of their neighbours. This method requires one pairwise communication transaction, but relies on the pairwise sharing of both robots' full states and the construction of the user-defined pseudomeasurements.
\subsection{Contributions and Paper Organization}
This paper addresses the problem of distributed filtering for collaborative localization, with the main contributions being:
\begin{itemize}
    \item a consensus-based distributed filtering framework which only shares \emph{relevant} shared states and RMIs over one pairwise communication exchange,
    \item a passive listening-based consensus update for neighbours that overhear the shared state exchange,
    \item validating the proposed approach in simulation and experiments, with results showing near centralized performance in consistency and accuracy compared to the state-of-the-art (SoTA) decentralized approach~\cite{cossette_decentralize_2024},
    \item open-source implementation to be released upon acceptance.
\end{itemize}

The remainder of this paper is organized as follows. Preliminary material is introduced in Section~\ref{sec:methodology}. The general problem statement is stated in Section~\ref{sec:problem_formulation}, followed by the presentation of the proposed consensus-based distributed filtering framework in Section~\ref{sec:consensus_based_estimation}. The simulation and experimental results are discussed in Section~\ref{sec:results}, and conclusions and future work are discussed in Section~\ref{sec:conclusion}.    
    \section{Methodology} \label{sec:methodology}
\subsection{Notation and Matrix Lie Groups}
Throughout this paper, a bold uppercase letter such as $\mbf{X} \in \mathbb{R}^{m \times n}$ denotes a matrix and a bold lowercase letter $\mbf{x}\in \mathbb{R}^n$ denotes a column matrix. 
Matrix Lie groups \SO{d} and \SE{d} represent rotations and poses in $d=\{2,3\}$, respectively. Another common MLG is the ``extended'' pose $SE_2(3)$ that encodes velocity information, in addition to position and attitude information~\cite{cossette_decentralize_2024}. 
The direction cosine matrix $\mbf{C}_{ab} \in \SO{d}$ enables a change of basis. That is, $\mbf{r}_a^{zw} = \mbf{C}_{ab} \mbf{r}_b^{zw}$ where $\mbf{r}_j^{zw} \in \mathbb{R}^d$ is the position of point $z$ relative to point $w$ resolved in reference frame $\rframe{j}$ where $j=\{a,b\}$.
If $G$ is an MLG, $\mathfrak{g}$ is the associated Lie algebra, where $\mathrm{exp}: \mathfrak{g} \rightarrow G$ and $\mathrm{log}: G \rightarrow \mathfrak{g}$ are the exponential and logarithm maps, respectively.
These mappings can be simplified~\cite{solà2021microlietheorystate} by introducing $\mathrm{Exp}:\mathbb{R}^m \to G$ and $\mathrm{Log} : G \to \mathbb{R}^m$, where for $\mc{X} \in G$ and $\mbs{\xi} \in \mathbb{R}^{m}$,
    $\mc{X} \triangleq \mathrm{Exp}(\mbs{\xi})$, and $\mbs{\xi} \triangleq \mathrm{Log}(\mc{X})$. 
A generalized addition $\oplus : G \times \mathbb{R}^m \to G$ and subtraction $\ominus : G \times G \to \mathbb{R}^m$ is also defined as 
    $\mc{X} = \bar{\mc{X}} \oplus \mbs{\xi}$, and $\mbs{\xi} = \mc{X} \ominus \bar{\mc{X}}$,
which can admit either a right or left perturbation definition as in~\cite{solà2021microlietheorystate}. The right-perturbation is used for all operations in this paper. The probability distribution of $\mc{X}$ is assumed to be a Gaussian distribution such that
$\mc{X} = \bar{\mc{X}}\oplus\delta\mbs{\xi}$ where $\bar{\mc{X}}$ is the mean group element and $\delta \mbs{\xi}\sim\mathcal{N}(\mbf{0}, \mbf{P})$ is a zero-mean perturbation in the Lie algebra with covariance $\mbf{P}$~\cite{bourmaud_intrinsic_2016}. For simplicity, the notation $p(\mc{X}) \triangleq \mathcal{N}_L(\bar{\mc{X}}, \mbf{P})$ is used to denote a Gaussian distribution of an MLG.
A composite group $\mbc{X}$ is defined as a concatenation of multiple Lie group states as $\mbc{X} = (\mc{X}_1, \dots, \mc{X}_N)$, where each $\mc{X}_i \in G_i$ with group operations occurring elementwise. 

\subsection{General Process and Measurement Models}
The general process model $\mbf{f}_i(\cdot)$ and measurement models $\mbf{g}_i(\cdot)$ that are local to a robot $i$ are 
\begin{align}
    \mc{X}_{i_{k}} &= \mbf{f}_i(\mc{X}_{i_{k-1}}, \mbf{u}_{{i}_{k-1}}, \mbf{w}_{i_{k-1}}), \hspace{-11pt} \quad &\mbf{w}_{i_{k-1}}\sim\mathcal{N}(\mbf{0}, \mbf{Q}_{{i}_{k-1}}), \label{eq:local_odom}\\
    \mbf{y}_{{i}_k} &= \mbf{g}_i(\mc{X}_{i_{k}}) + \mbf{v}_{i_{k}}, \hspace{-11pt} \quad &\mbf{v}_{i_{k}}\sim\mathcal{N}(\mbf{0}, \mbf{R}_{{i}_{k}})\label{eq:local_sensing}, 
\end{align}
where $\mc{X}_{i_{k}}$ is the state of robot $i$ at timestep $k$, which is dropped for brevity going forward. Here, $\mc{X}_{i}$ is a general Lie group element, but could also be a composite group. The local process input $\mbf{u}_{i}$ is from the robot's interoceptive sensor such as from a wheel odometer or an inertial measurement unit (IMU), and $\mbf{y}_{i}$ is a local observation from the robot's local exteroceptive sensor such as from GPS positioning or range measurements to known landmarks. Both the interoceptive and exteroceptive sensor measurements are corrupted with additive white noises, $\mbf{w}_{i}\sim\mathcal{N}(\mbf{0}, \mbf{Q}_{i})$ and
$\mbf{v}_{i}\sim\mathcal{N}(\mbf{0}, \mbf{R}_{i})$, respectively.
Robots are also capable of inter-robot sensing, such as from UWB inter-robot range measurements. The inter-robot measurement model $\mbf{h}_{ij}(\cdot)$ is a function of both robots $i$ and $j$'s states, that is 
\begin{align}
    \mbf{y}_{ij_{k}} = \mbf{h}_{ij}(\mc{X}_{i_{k}}, \mc{X}_{j_{k}}) + \mbf{n}_{ij_{k}}, \quad \mbf{n}_{ij_{k}}\sim\mathcal{N}(\mbf{0}, \mbf{R}_{ij_{k}}), \label{eq:inter_robot_measurement}
\end{align}
where $\mbf{y}_{ij}$ is the inter-robot observation between robots $i$ and $j$, that is corrupted by additive white noise $\mbf{n}_{ij}\sim\mc{N}(\mbf{0}, \mbf{R}_{ij})$. The inter-robot measurement can be available to either robot $i$, robot $j$, or both. 

\subsection{Centralized vs Distributed Estimators}
The definition of a centralized and distributed estimator for a team of $N$ robots are as follows.

\subsubsection{Centralized Estimators}
Centralized estimators attempt to estimate the global state $\mbc{X} = (\mc{X}_{1}, \dots, \mc{X}_{N})$ where $\mc{X}_i \in G_i$, with access to all local measurements $\mbf{u}_i, \mbf{y}_i$, and inter-robot measurements $\mbf{y}_{ij}$ for all robots $i, j \in \{1, \dots, N\}$ at an arbitrary timestep. By fusing all available sensor information, centralized estimators are theoretically optimal in attaining the lowest estimation error covariance, but are rarely implemented in practice due to their demanding communication and infrastructure requirements.

\subsubsection{Distributed Estimators}
Distributed estimators are local to all $N$ robots and attempt to estimate a subset of the global state $\mbc{X}_i \subseteq \mbc{X}$, for a robot $i$. In this work, a robot $i$ estimates its local state $\mc{X}_{i;i}$ using its local measurements $\mbf{u}_i, \mbf{y}_i$, \emph{and} the states of its one-hop neighbours $\mc{X}_{j;i}, \forall j \in \mathcal{N}_i$ using inter-robot measurements $\mbf{y}_{ij}$ and information received from robot $j$, where $\mc{N}_i$ is the set of one-hop neighbours of robot $i$. The full state of robot $i$ is thus
\begin{align}
    \mbc{X}_i = (\mc{X}_{i;i}, \{\mc{X}_{j;i}\}_{j\in\mathcal{N}_i}),
\end{align}
where the subscript notation $(\cdot)_{;i}$ indicates that the state $(\cdot)$ is estimated by robot $i$.
A \emph{shared state} between robots $i$ and $j$ is the intersection of their full states $\mbc{X}_i \cap \mbc{X}_j$, which includes shared neighbours. 
For example, consider the communication graph for $N=4$ robots in Fig.~\ref{fig:comm_graph}, with global state $\mbc{X} = (\mc{X}_1, \mc{X}_2, \mc{X}_3, \mc{X}_4)$. The full state of Robot 1 is $\mbc{X}_1 = (\mc{X}_{1}, \mc{X}_{2}, \mc{X}_{3}, \mc{X}_{4})_{;1}$, and the full state of Robot 2 is $\mbc{X}_2 = (\mc{X}_{1}, \mc{X}_{2}, \mc{X}_{4})_{;2}$. Hence, the shared state between Robot 1 and Robot 2 as estimated by Robot 1 is $\mbc{X}_{1_{(1,2)}} = (\mc{X}_{1}, \mc{X}_{2}, \mc{X}_{4})_{;1}$. Thus, the full state of Robot 1 can be partitioned into the shared state $\mbc{X}_{1_{(1,2)}}$ and the unique state $\mc{X}_{3;1}$ as $\mbc{X}_1 = (\mbc{X}_{1_{(1,2)}}, \mc{X}_{3;1})$. This notation and partitioning can be generalized to any pair of robots $i$ and $j$. The benefit of this will be apparent in Section~\ref{sec:consensus} when forming the consensus distribution of the shared states.

\subsection{Preintegrated Odometry Sharing}
For a robot $i$ to propagate its estimate of a neighbour $j$'s state $\mc{X}_{j;i}$, it must have access to the odometry information of robot $j$. As in~\cite{shalaby_multi_robot_2024, cossette_decentralize_2024}, robot $i$ holds a preintegrated process model $\mbf{f}_{pq}(\cdot)$ relating the neighbour state $\mc{X}_{j;i}$ at two arbitrary timestamps $k=p$ and $k=q$ as
\begin{align}
    \hspace{-5pt} \mc{X}_{j;i_{q}} = \mbf{f}_{pq}(\mc{X}_{j;i_{p}}, \Delta \mc{X}_{j_{pq}}) \oplus \mbf{w}_{pq}, \quad \mbf{w}_{pq}\sim\mathcal{N}(\mbf{0}, \mbf{Q}_{pq}), \label{eq:neigh_odom}
\end{align}
where $\Delta \mc{X}_{j_{pq}}$ is the preintegrated odometry, or RMI, and $\mbf{Q}_{pq}$ is the preintegrated noise covariance, both received from robot $j$. When no RMI is available, the neighbour's state is propagated with a ``zero'' RMI. This allows for a constant-memory, on-demand sharing of odometry information between two arbitrarily long timestamps, without needing to share each individual high-rate odometry measurement.
   
    \section{Problem Formulation}\label{sec:problem_formulation}
Consider a team of $N$ robots that form an undirected communication graph $\mathcal{G}=(\mathcal{V}, \mathcal{E})$ where $\mathcal{V}=\{1, 2, \ldots, N\}$ is the set of vertices representing the robots and $\mathcal{E}$ is the set of edges representing communication links between robots. For an arbitrary robot $i$, the goal is to estimate its full state $\mbc{X}_{i_{k}} = (\mbf{T}_{ai;i_{k}}, \{\mbf{T}_{aj;i_{k}}\}_{j\in\mc{N}_i})$, comprised of its local (extended) pose $\mbf{T}_{ai;i_{k}}$ and neighbouring poses $\{\mbf{T}_{aj;i_{k}}\}_{j\in\mc{N}_i}$, all relative to the inertial reference frame $\rframe{a}$. The local pose is propagated through local odometry, local sensing, and inter-robot measurements, while the neighbouring poses are propagated by receiving neighbouring RMIs and inter-robot measurements. Both the local and neighbouring states are also updated through either direct (Section~\ref{sec:direct_consensus}) or passive (Section~\ref{sec:passive_consensus}) consensus updates. The initialization procedure is assumed to be known, such that all robots hold the same initial beliefs for their shared states. It is assumed that only pairwise communication is available to minimize communication interference. It is also assumed that RMIs can be passively received by any robot that is in range of either an inter-robot measurement or a consensus exchange. Finally, the communication topology is static, with a fixed, known communication schedule, where robots are able to extract the relevant shared state information during a consensus exchange.
   
    \section{Consensus-Based Distributed Estimation} \label{sec:consensus_based_estimation}
This section will describe updating the full state based on first forming a consensus distribution of the shared states between a pair of robots $i$ and $j$, which is then used to update their unique states. The motive and goal of cyclically forming a consensus distribution among pairs of the network is for all full states to converge onto the centralized solution.
\subsection{Consensus Distribution of Shared States}\label{sec:consensus}
Along with sharing RMIs, it is assumed that robots can also pairwise exchange their shared state estimates. For a pair of robots $i$ and $j$ that estimate the same physical quantities, the shared states are in \emph{consensus} if 
\begin{align}
    \mbc{X}_{i_{(i,j)}} = \mbc{X}_{j_{(i,j)}}. \label{eq:consensus}
\end{align}
From robot $i$'s perspective prior to consensus, its marginal density of the full state $p(\mbc{X}_i)$ can be partitioned into the joint density of the shared and unique components as
\begin{align}
    p(\mbc{X}_{i_{S}}, \mbc{X}_{i_{U}}) \sim \mathcal{N}_L\left(\begin{bmatrix}\hat{\mbc{X}}_{i_{S}} \\ \hat{\mbc{X}}_{i_{U}}\end{bmatrix}, \begin{bmatrix}\mbfhat{P}_{i_{SS}} & \mbfhat{P}_{i_{SU}} \\ \mbfhat{P}_{i_{US}} & \mbfhat{P}_{i_{UU}}\end{bmatrix}\right),
\end{align} 
where $S$ and $U$ subscripts denote the shared and unique components, respectively. By Bayes' theorem, the above joint density can be factored into the marginal density of the shared states and the conditional density of the unique states given the current shared state estimate as~\cite[Sec.~2.2.2]{Barfoot_2024}
\begin{align}
    p(\mbc{X}_{i_{S}}, \mbc{X}_{i_{U}}) & = p(\mbc{X}_{i_{S}})p(\mbc{X}_{i_{U}}|\mbc{X}_{i_{S}}), \label{eq:joint_density} \\
    p(\mbc{X}_{i_{S}}) &\sim \mathcal{N}_L(\hat{\mbc{X}}_{i_{S}}, \mbfhat{P}_{i_{SS}}),\\
    p(\mbc{X}_{i_{U}}|\mbc{X}_{i_{S}}) &\sim \mathcal{N}_L(\tilde{\mbc{X}}_{i_{U}}, \tilde{\mbf{P}}_{i_{UU}}), \label{eq:conditional_dist}\\
    \tilde{\mbc{X}}_{i_{U}} &= \hat{\mbc{X}}_{i_{U}} \oplus \mbf{M}(\mbc{X}_{i_{S}} \ominus \hat{\mbc{X}}_{i_{S}}), \\
    \tilde{\mbf{P}}_{i_{UU}} &= \mbfhat{P}_{i_{UU}} - \mbf{M}\mbfhat{P}_{i_{SU}}, \quad \mbf{M} = \mbfhat{P}_{i_{US}}\mbfhat{P}_{i_{SS}}^{-1}, 
\end{align}
where the generalized addition and subtraction have overloaded the typical vector space definition. To drive consensus between the shared states, robots $i$ and $j$ share their local belief of the shared states $(\hat{\mbc{X}}_{i_{S}}, \mbfhat{P}_{i_{S}})$ and $(\hat{\mbc{X}}_{j_{S}}, \mbfhat{P}_{j_{S}})$ to eachother, respectively. Hence, both robots $i$ and $j$ have access to both beliefs of the shared states.
Note that inter-robot estimates are correlated due to double counting RMIs and possibly sharing inter-robot measurements of the form~\eqref{eq:inter_robot_measurement}. This is addressed by first performing a CI step between the shared states' covariances locally on robot $i$ as
    $\mbfhat{P}_{i_{S}} \leftarrow \f{1}{\omega_i}\mbfhat{P}_{i_{S}}$, $\mbfhat{P}_{j_{S}} \leftarrow \f{1}{1-\omega_i}\mbfhat{P}_{j_{S}}$
where $\omega_i \in [0,1]$ is robot $i$'s CI weight that is handpicked, or optimized by minimizing the trace or log-determinant of the fused covariance~\cite{julier_non-divergent_1997}. From the perspective of robot $j$, the subscripts are swapped as $\omega_j = 1-\omega_i$.

As the shared poses are states on MLGs, their fusion is performed as in~\cite{barfoot_associating_2014}. From the perspective of robot $i$, the procedure of forming the consensus distribution of the shared states between $i$ and $j$ begins with the CI step. Now, robot $i$ has access to both $(\hat{\mbc{X}}_{i_{S}}, \mbfhat{P}_{i_{S}})$ and $(\hat{\mbc{X}}_{j_{S}}, \mbfhat{P}_{j_{S}})$ noting that the covariances are inflated due to CI. The goal is to form an optimization problem to solve for the optimal shared state $\mbc{X}_{S}$, that can be executed on both robots $i$ and $j$.
The residual that is sought to be minimized is formed as $\mbf{e}_i(\mbc{X}_{S}) = \hat{\mbc{X}}_{i_{S}} \ominus \mbc{X}_{S}$. Further expanding $\mbc{X}_{S}$ into nominal and perturbation terms as $\mbc{X}_{S} = \mbc{X}_{\mathrm{op}} \oplus \mbs{\epsilon}$, the residual can be rewritten in terms of the perturbation $\mbs{\epsilon}$ as 
\begin{align}
    \mbf{e}_i(\mbc{X}_{S}) = \hat{\mbc{X}}_{i_{S}} \ominus (\mbc{X}_{\mathrm{op}} \oplus \mbs{\epsilon}).
\end{align} 
With the assumption that the perturbation $\mbs{\epsilon}$ is small, by the Baker-Campbell-Hausdorff (BCH) approximation~\cite[Sec.~8.1.5]{Barfoot_2024}, the fusion of MLG states is reformed as a weighted nonlinear least-squares problem of the form~\cite{barfoot_associating_2014}
\begin{align}
    \sum_{n}^{(i,j)}\left(\mbf{G}_n^\trans\mbf{P}_{n_{S}}^{-1}\mbf{G}_n \right)\mbs{\epsilon} = \sum_{n}^{(i,j)}\mbf{G}_n^\trans\mbf{P}_{n_{S}}^{-1}\mbf{e}_n(\mbc{X}_{\mathrm{op}}).
\end{align} 
Here, $\mbf{G}_n$ is the (left or right, depending on chosen perturbation direction) inverse group Jacobian evaluated at $\mbf{e}_n(\mbc{X}_{\mathrm{op}})$. The problem is iteratively updated by Gauss-Newton with $\mbc{X}_{\mathrm{op}} \leftarrow \mbc{X}_{\mathrm{op}} \oplus \mbs{\epsilon}$ until a convergence criterion is met. The consensus distribution is formed at convergence as 
\begin{align}
    \hat{\mbc{X}}_{S}^* &= \mbc{X}_{\mathrm{op}}, \qquad \mbfhat{P}_{S}^* = (\mbf{G}_i^\trans\mbf{P}_{i_{S}}^{-1}\mbf{G}_i + \mbf{G}_j^\trans\mbf{P}_{j_{S}}^{-1}\mbf{G}_j)^{-1}, \label{eq:prop_consensus} 
\end{align} 
where $p(\mbc{X}^*_{S}) \sim \mathcal{N}(\hat{\mbc{X}}_{S}^*, \mbfhat{P}_{S}^*)$ on both robots $i$ and $j$. In this paper, from the perspective of robot $i$, $\mbc{X}_{\mathrm{op}}$ is initialized as
\begin{align}
    \mbc{X}_{\mathrm{op}} = \hat{\mbc{X}}_{i_{S}} \oplus (\alpha (\hat{\mbc{X}}_{i_{S}}\ominus \hat{\mbc{X}}_{j_{S}})), \label{eq:initialization}
\end{align}
where $\alpha$ serves to perturb the local shared state estimate towards the neighbour's shared state estimate. For simplicity, $\alpha=0.5$ to give equal weight to both estimates, but the CI weight $\omega_i$ could also be used to weigh by uncertainty.
\subsection{Updating the Local Joint Distribution with Consensus}\label{sec:direct_consensus}
Herein lies a novel contribution of the proposed method in reconditioning the local joint distribution using the newly formed consensus distribution of the shared states. From the perspective of robot $i$, the local joint density is updated by fusing the newly formed marginal consensus distribution $p(\mbc{X}^*_{S})$ and the conditional distribution~\eqref{eq:conditional_dist} as
\begin{align}
    p(\mbc{X}_i^*) & = p(\mbc{X}_{S}^*, \mbc{X}_{i_{U}}^*) = p(\mbc{X}_{{S}}^*)p(\mbc{X}_{i_{U}}|\mbc{X}_{S}^*).
\end{align}
The random variables representing the shared and unique states of robot $i$ can be expressed as 
\begin{align}
    \mbc{X}_{S}^* & = \hat{\mbc{X}}_{S}^* \oplus \delta\mbs{\xi}_S, \quad \delta\mbs{\xi}_S \sim \mathcal{N}(\mbf{0}, \mbfhat{P}_{S}^*), \\
    \mbc{X}_{i_{U}}^* & = \hat{\mbc{X}}_{i_{U}} \oplus (\mbf{M}\underbrace{((\hat{\mbc{X}}_{{S}}^* \oplus \delta\mbs{\xi}_S) \ominus \hat{\mbc{X}}_{i_{S}})}_{\delta \mbs{\xi}_S^*} + \delta\mbs{\xi}_U),
\end{align}
where $\delta\mbs{\xi}_U \sim \mathcal{N}(\mbf{0}, \tilde{\mbf{P}}_{i_{UU}})$. Note that here the conditional noise $\delta\mbs{\xi}_U$ is added to the consensus noise term $\mbf{M}\delta\mbs{\xi}_S^*$ in the tangent space of the unique states.
The means of the marginal consensus distribution and the conditional distributions are 
\begin{align}
     \mathbb{E}[\mbc{X}^*_S] = \hat{\mbc{X}}_{S}^*, \quad
     \bar{\mbc{X}}_{i_{U}} \triangleq \mathbb{E}[\mbc{X}_{i_{U}}^*]  = \hat{\mbc{X}}_{i_{U}} \oplus \mbf{M}\delta\mbfbar{x},
\end{align}
where $\delta\mbfbar{x} = \hat{\mbc{X}}_{{S}}^* \ominus \hat{\mbc{X}}_{i_{S}}$. When updating a covariance component that includes the unique states, the term
\begin{align}
    \mbc{X}_{i_{U}}^* \ominus \bar{\mbc{X}}_{i_{U}} & = (\hat{\mbc{X}}_{i_{U}} \oplus (\mbf{M}\delta\mbs{\xi}_S^* + \delta\mbs{\xi}_U)) \ominus \bar{\mbc{X}}_{i_{U}}, \label{eq:exp_unique}
\end{align}
must be computed. As perturbations are small, $\delta \mbs{\xi}^*_S$ is approximated using the BCH approximation as
\begin{align}
\delta \mbs{\xi}^*_S = (\hat{\mbc{X}}_{{S}}^* \oplus \delta\mbs{\xi}_S) \ominus \hat{\mbc{X}}_{i_{S}} \approx \delta\mbfbar{x} + \mbc{J}_r^{-1}(\delta\mbfbar{x})\delta\mbs{\xi}_S, \label{eq:BCH_approx_recond} 
\end{align}
where $\mbc{J}_r(\cdot)$ is the right group Jacobian of the MLG.
When substituting~\eqref{eq:BCH_approx_recond} into~\eqref{eq:exp_unique}, the term 
\begin{align}
    \mathrm{Exp}(\mbf{M}\delta\mbfbar{x} + \mbf{M}\mbc{J}_r^{-1}(\delta\mbfbar{x})\delta\mbs{\xi}_S + \delta\mbs{\xi}_U)  \label{eqn:approx_1_BCH}
\end{align}
emerges and can be further approximated by the BCH approximation as 
\begin{align}
    &\mathrm{Exp}(\mbf{M}\delta\mbfbar{x} + \mbf{M}\mbc{J}_r^{-1}(\delta\mbfbar{x})\delta\mbs{\xi}_S + \delta\mbs{\xi}_U) \notag\\
    \hspace{-6.5pt}\approx &\mathrm{Exp}(\mbf{M}\delta\mbfbar{x})\mathrm{Exp}(\mbc{J}_r(\mbf{M}\delta\mbfbar{x})(\mbf{M}\mbc{J}_r^{-1}(\delta\mbfbar{x})\delta\mbs{\xi}_S + \delta\mbs{\xi}_U)). \label{eqn:approx_2_BCH}
\end{align}
After substituting~\eqref{eqn:approx_2_BCH} back into~\eqref{eq:exp_unique} and simplifying, the final approximation can be retrieved as
\begin{align}
    \mbc{X}_{i_{U}}^* \ominus \bar{\mbc{X}}_{i_{U}} & \approx \mbf{J}_S\delta\mbs{\xi}_S + \mbf{J}_U\delta\mbs{\xi}_U,
\end{align}
where $\mbf{J}_S = \mbc{J}_r(\mbf{M}\delta\mbfbar{x})\mbf{M}\mbc{J}_r^{-1}(\delta\mbfbar{x})$ and $\mbf{J}_U = \mbc{J}_r(\mbf{M}\delta\mbfbar{x})$.
Now, assuming $\delta\mbs{\xi}_S$ and $\delta\mbs{\xi}_U$ are independent, the reconditioned unique states and covariances of robot $i$ are 
\begin{align}
    \hspace{-5pt}\hat{\mbc{X}}_i = \begin{bmatrix}\hat{\mbc{X}}_{S}^* \\ \bar{\mbc{X}}_{i_{U}}\end{bmatrix}, \: \mbfhat{P}_i = \begin{bmatrix}\mbfhat{P}_{S}^* & \mbfhat{P}_{S}^*\mbf{J}_S^\trans \\ \mbf{J}_S\mbfhat{P}_{S}^* & \mbf{J}_S\mbfhat{P}_{S}^*\mbf{J}_S^\trans + \mbf{J}_U\mbftilde{P}_{i_{UU}}\mbf{J}_U^\trans\end{bmatrix}.
\end{align} 
If the local shared state $\hat{\mbc{X}}_{i_{S}}$ and the consensus shared state $\hat{\mbc{X}}_{i_{S}}^*$ are equal, there will be no change in the unique component of the local estimate. That is, if $\delta\mbfbar{x}\approx \mbf{0}$, $\mbc{J}_r(\cdot) \approx \eye$ the updated covariance can be further approximated as
\begin{align}
    \mbfhat{P}_i & \approx \begin{bmatrix}\mbfhat{P}_{S}^* & \mbfhat{P}_{S}^* \mbf{M}^\trans \\ \mbf{M}\mbfhat{P}_{S}^* & \mbf{M}\mbfhat{P}_{S}^*\mbf{M}^\trans + \tilde{\mbf{P}}_{i_{UU}}\end{bmatrix}.
\end{align}
\subsection{Passive Listening on Shared State Exchange}\label{sec:passive_consensus}
Motivated by the passive listening approach, another novel contribution of the proposed method is to allow non-communicating robots (herein passive robots) to passively listen in on the shared state exchange between communicating robots. In doing so, passive robots can update their local belief on the subset of broadcasted shared states that they also estimate. For example, in Fig.~\ref{fig:comm_graph}, Robot 3 and Robot 4 are passive robots that have full state $\mbc{X}_3=(\mc{X}_1, \mc{X}_3, \mc{X}_4)_{;3}$ and $\mbc{X}_4=(\mc{X}_1, \mc{X}_2, \mc{X}_3, \mc{X}_4)_{;4}$, respectively. When Robot 1 and Robot 2 broadcast their shared states $\mbc{X}_{1_{(1,2)}}$ and $\mbc{X}_{2_{(1,2)}}$, the passive robots can update their relevant shared state estimates locally as per the proposed method described in Section~\ref{sec:consensus} and Section~\ref{sec:direct_consensus}. This can benefit the passive robots if the communication schedule is long and infrequent, as they can still benefit from the information being communicated.   
    \section{Simulation and Experimental Results} \label{sec:results}
The proposed method outlined in Section~\ref{sec:consensus_based_estimation} is evaluated using both simulated and experimental data. In all simulations and experiments, the results are compared to the centralized estimator as a baseline, and the SoTA decentralized approach proposed in~\cite{cossette_decentralize_2024}. When implementing~\cite{cossette_decentralize_2024}, a consensus-style pseudomeasurement like~\eqref{eq:consensus} is used, with handset CI weight $\omega = 0.99$ and $\mbs{\Psi} = \mbf{0}$. For simulation results, the proposed method solves for $\omega$ dynamically by minimizing the trace of the fused covariance matrix, whereas it is fixed at $\omega = 0.5$ for all experimental results to maintain a static comparison. Unless specified otherwise, the proposed and SoTA methods use the passive listening approach described in Section~\ref{sec:passive_consensus}. All methods use an on-manifold extended Kalman filter for state estimation~\cite{solà2021microlietheorystate}, where outlier measurements are rejected by the \emph{normalized-innovation-squared} test~\cite[Sec.~5.1.2]{Barfoot_2024}. 

\subsection{Evaluation Metrics}
All estimators are evaluated based on accuracy and consistency. The distributed estimators are also evaluated on how closely they match the centralized estimates.
\begin{enumerate}[a)]
    \item Accuracy: The accuracy of the estimators are evaluated based on the root mean squared error (RMSE) of the estimates, $\hat{\mbc{X}}$, compared to the groundtruth, $\mbc{X}$. For $M$ total Monte Carlo (MC) trials, the estimation error for trial $m$ at timestep $k$ is computed as $\mbf{e}_{m,k} = \mbc{X}_{m,k} \ominus \hat{\mbc{X}}_{m,k}$, the (per-component) $M$-trial averaged RMSE and temporally averaged RMSE ($\overline{\textrm{aRMSE}}$) are computed as 
    $\textrm{aRMSE}_k = \sqrt{\f{1}{M}\sum_{m=1}^{M}\f{1}{n} \mbf{e}_{m,k}^\trans\mbf{e}_{m,k}}$ and $\overline{\textrm{aRMSE}} = \f{1}{K+1}\sum_{k=0}^K \textrm{aRMSE}_k$,
    for $n$ component (position, attitude, etc.) degrees of freedom and $K+1$ total timesteps.
    \item Consistency: A consistent estimator is one that produces covariance estimates that properly captures the true uncertainty of the estimation errors. The consistency of the estimators are qualitatively evaluated based on the $\pm 3\sigma$ bounds enveloping the estimation errors, and quantitatively evaluated based on the temporally averaged normalized estimation error squared (NEES)~\cite[Sec.~5.4]{barshalom_estimation_2002}, which is computed as
        $\textrm{NEES} = \f{1}{(K+1)n}\sum_{k=0}^K\mbf{e}_{k}^\trans\mbs{\Sigma}_{k}^{-1}\mbf{e}_{k}$,
    where a consistent estimator should have a NEES close to one.
    \item Comparability to Centralized Estimates: The distributed estimators are compared to the relevant centralized estimates based on the 2-Wasserstein distance between their distributions~\cite[Sec.~2.6]{peyré2020computationaloptimaltransport}. For Gaussian distributions, the (squared) 2-Wasserstein distance between $\mbf{x}\sim \mc{N}(\mbs{\mu}_x, \mbs{\Sigma}_x)$ and $\mbf{y}\sim \mc{N}(\mbs{\mu}_y, \mbs{\Sigma}_y)$ is 
    $\mathcal{W}_2({\mbf{x}},\mbf{y})^2 = \norm{\mbs{\mu}_x \ominus \mbs{\mu}_y}_2^2 + \mathcal{B}(\mbs{\Sigma}_{x}, \mbs{\Sigma}_{y})^2$,
    where the normal vector space definition has been overloaded. The first term is the (squared) Euclidean distance between both distributions' means and the second term is the (squared) \emph{Bures} metric between their covariances. Explicitly the (squared) \emph{Bures} metric is
    $\mathcal{B}(\mbs{\Sigma}_{x}, \mbs{\Sigma}_{y})^2 = \textrm{tr}(\mbs{\Sigma}_x + \mbs{\Sigma}_y - 2(\mbs{\Sigma}_y^{\f{1}{2}}\mbs{\Sigma}_x \mbs{\Sigma}_y^{\f{1}{2}})^{\f{1}{2}})$. 
\end{enumerate}

\subsection{Simulation Results}
The proposed method is first validated in simulation, which considers $N$ robots moving in 2D. The robots are equipped with gyroscopes and wheel odometers, generating angular and forward body frame velocity measurements. All robots are also equipped with two UWB transceivers at known lever arms, generating attitude-coupled inter-robot range measurements~\cite{shalaby_relative_2021}. Some robots are equipped with GPS, generating global position measurements. It is only through inter-robot communication that non-GPS-equipped robots can estimate their global poses. 
For all simulations, odometry measurements occur at $100~\si{Hz}$, all-to-all UWB-range measurements at $50~\si{Hz}$, and GPS measurements at $10~\si{Hz}$. Regardless of the size of the communication schedule, it is cycled through at $10~\si{Hz}$. The following simulation scenarios are considered.
\begin{enumerate}
    \item S1: $N\in\{3, 5, 9, 17\}$, where only Robot 1 is equipped with GPS. The communication graph is sparse, where only Robot 1 can communicate with all other robots.
    \item S2: $N=4$, where Robots 1 and 4 are equipped with GPS. The communication graph is equivalent to the one shown in Fig.~\ref{fig:comm_graph}.
\end{enumerate}
\begin{figure*}[!t]
    \centering
    \begin{subfigure}{0.496\textwidth}
        \centering
        \includegraphics[width=\linewidth]{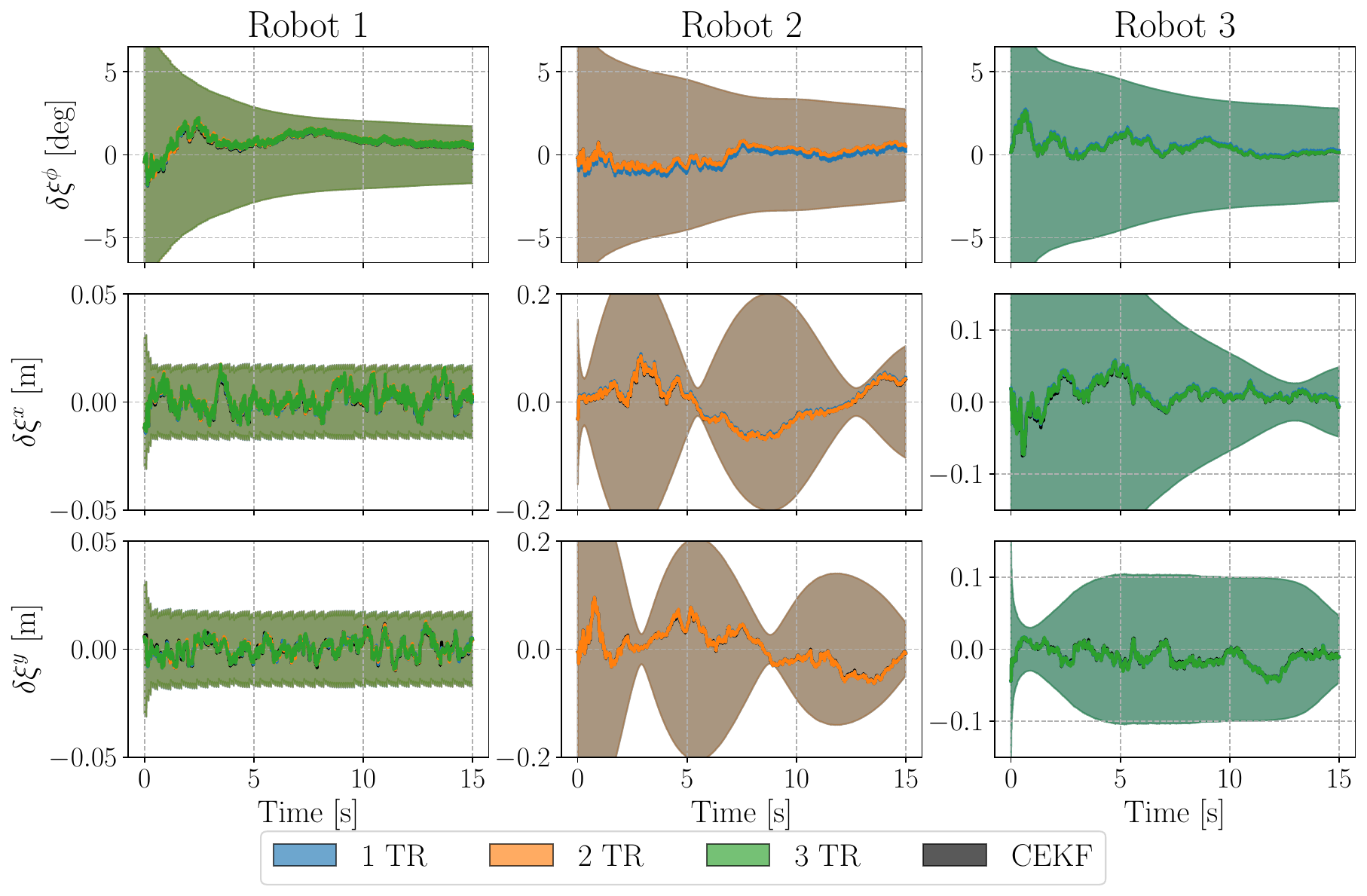}
        \caption{Proposed Method}
        \label{fig:error_S1_N3_prop}
    \end{subfigure}
    \hfill
    \begin{subfigure}{0.496\textwidth}
        \centering
        \includegraphics[width=\linewidth]{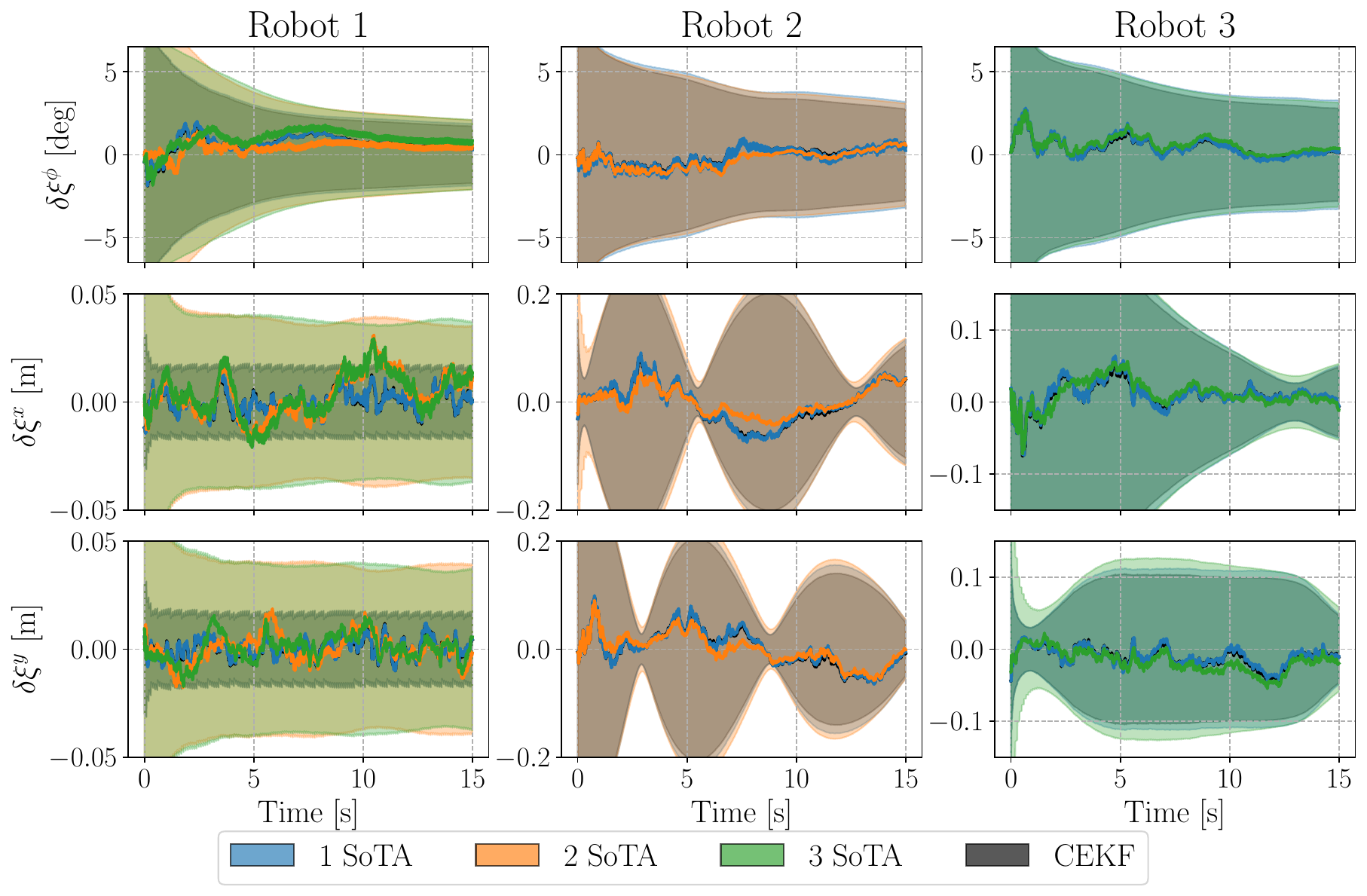}
        \caption{SoTA Method}
        \label{fig:error_S1_N3_sota}
    \end{subfigure}
    \caption{Error plots and $\pm3\sigma$ bounds (shaded region) of all distributed estimators for S1 with $N=3$ robots, comparing (a) proposed and (b) SoTA methods. Subplots share the same axis limits for a qualitative comparison.}
    \vspace{-3pt}
    \label{fig:error_S1_N3}
\end{figure*}
\begin{figure}[h]
    \centering
    \includegraphics[width=\linewidth]{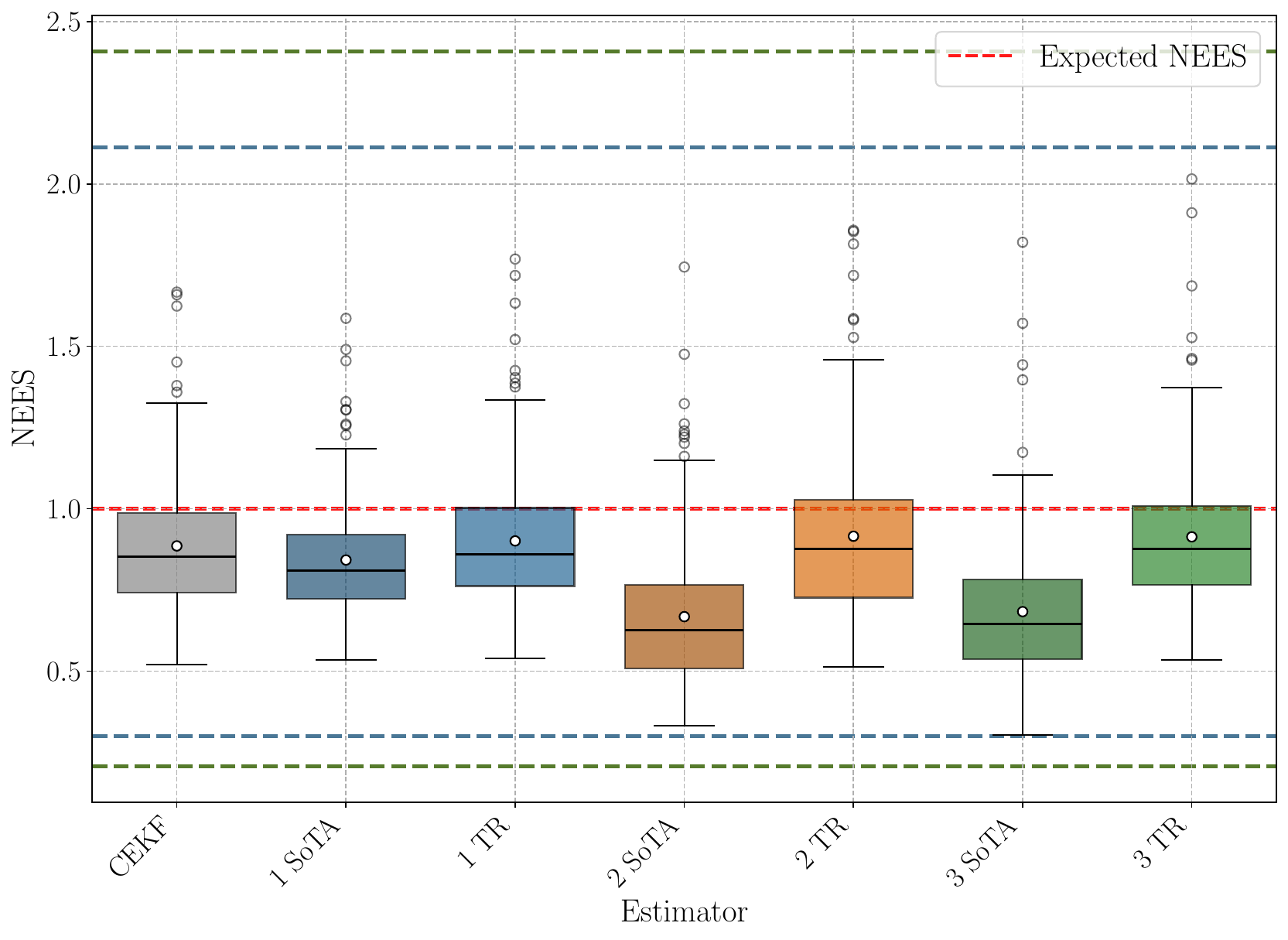}
    \caption{Comparison of the NEES between proposed (TR) and SoTA~\cite{cossette_decentralize_2024} estimators for S1, with $N=3$ over $M = 200$ MC runs. The white dot indicates the mean whereas the black line indicates the median.} 
    \vspace{-3pt}
    \label{fig:anees_comparison_S1}
\end{figure}
\begin{figure}[h]
    \centering
    \includegraphics[width=\linewidth]{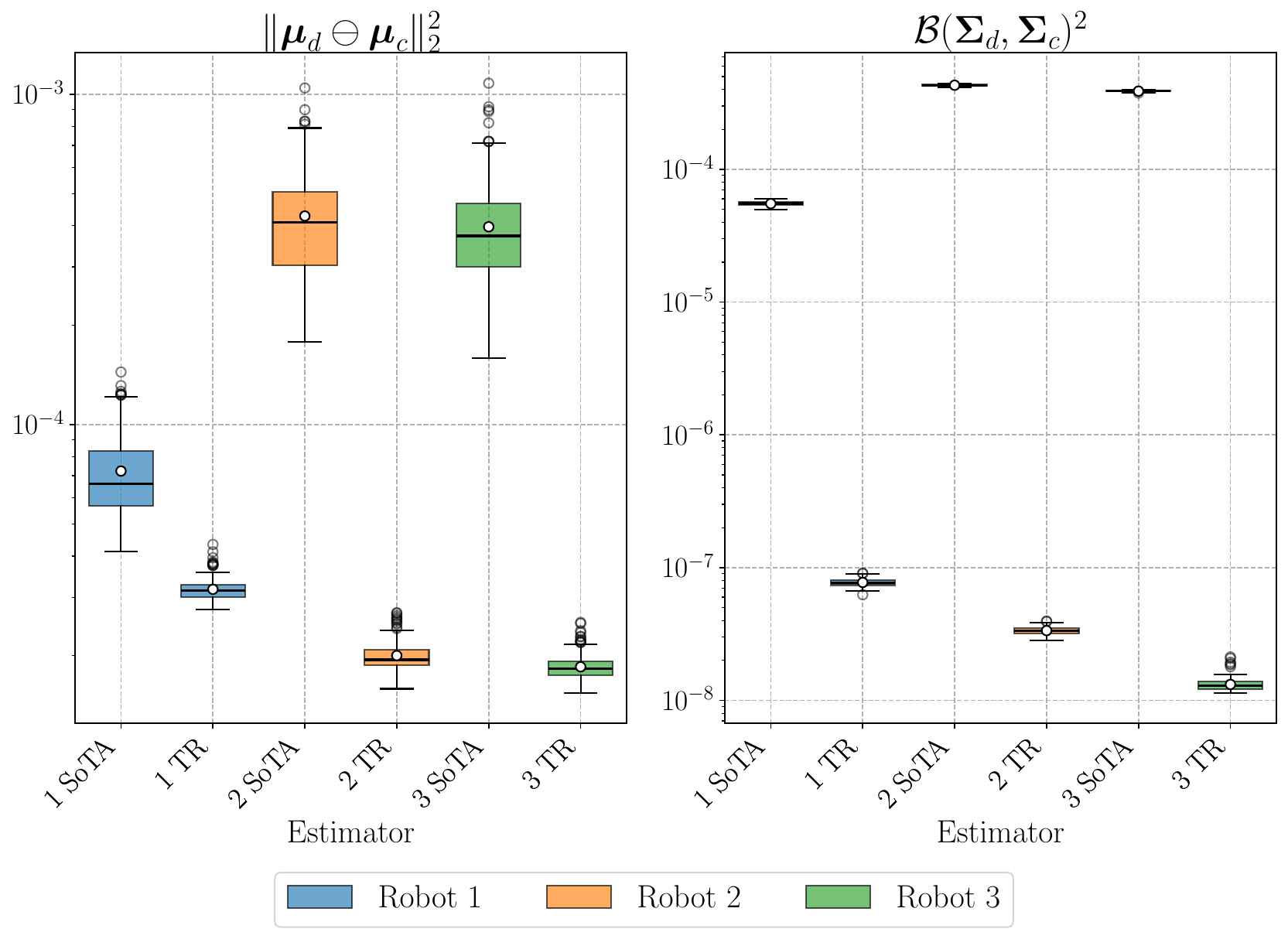}
    \caption{Component-wise comparison of the 2-Wasserstein metrics between proposed (TR) and SoTA~\cite{cossette_decentralize_2024} estimators for S1, with $N=3$ over $M = 200$ MC runs.} 
    \vspace{-3pt}
    \label{fig:wasserstein_comparison_S1}
\end{figure}
The $\pm3\sigma$ error plots for one trial of S1 with $N=3$ is shown in Fig.~\ref{fig:error_S1_N3}, comparing both methods. While both solutions are accurate, it is clear that the proposed method is more precise in the uncertainty estimate of the neighbouring states, particularly in Robot 2 and 3's estimate of Robot 1. Moreover, the distributed estimates of the proposed approach are in consensus as the errors and shaded $\pm3\sigma$ regions are tightly overlapping, nearing the centralized solution. This increase in consistency is further shown in the NEES boxplot in Fig.~\ref{fig:anees_comparison_S1}, where all estimators of the proposed method are more consistent than the SoTA method, also nearing the centralized solution. Fig.~\ref{fig:wasserstein_comparison_S1} shows the temporally averaged component-wise 2-Wasserstein metrics, and it is clear that both the mean and covariance estimates of the proposed method are closer to the centralized solution compared to the SoTA method.
\begin{table}[h]
    \footnotesize
    \centering
    \caption{Position $\overline{\textrm{aRMSE}}$ [\si{m}] of Robot 1 as estimated by itself, and its neighbours, with and without passive listening. Percentages show relative difference from the self-estimate.}
    \begin{tabular}{c*{4}{c}}
        \toprule
        & \multicolumn{2}{c|}{Self} & \multicolumn{2}{c}{Averaged Neighbour Value} \\
        \cmidrule(lr){2-5}
        \makecell{Number of\\Neighbours} & w/o Psv. & w/ Psv.  & w/o Psv. & w/ Psv.  \\
        \midrule
        2 & 0.0054 & 0.0054 & 0.0061 (13.0\%) & 0.0054 (0.0\%) \\        
        4 & 0.0053 & 0.0053 & 0.0070 (32.1\%) & 0.0054 (1.9\%)   \\    
        8 & 0.0051 & 0.0051 & 0.0085 (66.7\%) & 0.0053 (3.9\%)  \\
        16& 0.0049 & 0.0049 & 0.0108 (120.4\%) & 0.0051 (4.1\%)  \\        
        \bottomrule
    \end{tabular}
    \label{tab:sim_rmse_comparison_1}
    \vspace{-3pt}
\end{table}
To show the benefit of passive listening, Table~\ref{tab:sim_rmse_comparison_1} compares the positioning $\overline{\textrm{aRMSE}}$ of Robot 1's position as estimated by Robot 1 and the average of all its neighbours' estimates. Without passive listening, neighbour estimates must wait the entire communication schedule to perform a consensus update, which results in longer delays as the teamsize increases. During this time, neighbours can only deadreckon their belief of Robot 1's state using received RMIs, indicated by the increasing averaged neighbour $\overline{\textrm{aRMSE}}$ with increasing team size. However, when passive listening is enabled, neighbours can perform passive consensus updates more frequently, which results in more accurate estimates of Robot 1's state even with increasing team size. It should also be observed that the neighbours' accuracy tends to converge to the self-estimate of Robot 1, indicating that the neighbours are in consensus with Robot 1's estimate. For this simulation, passive listening does not impact the accuracy of Robot 1's self-estimate due to the communication topology.
\begin{figure}[h]
    \centering
    \includegraphics[width=\linewidth]{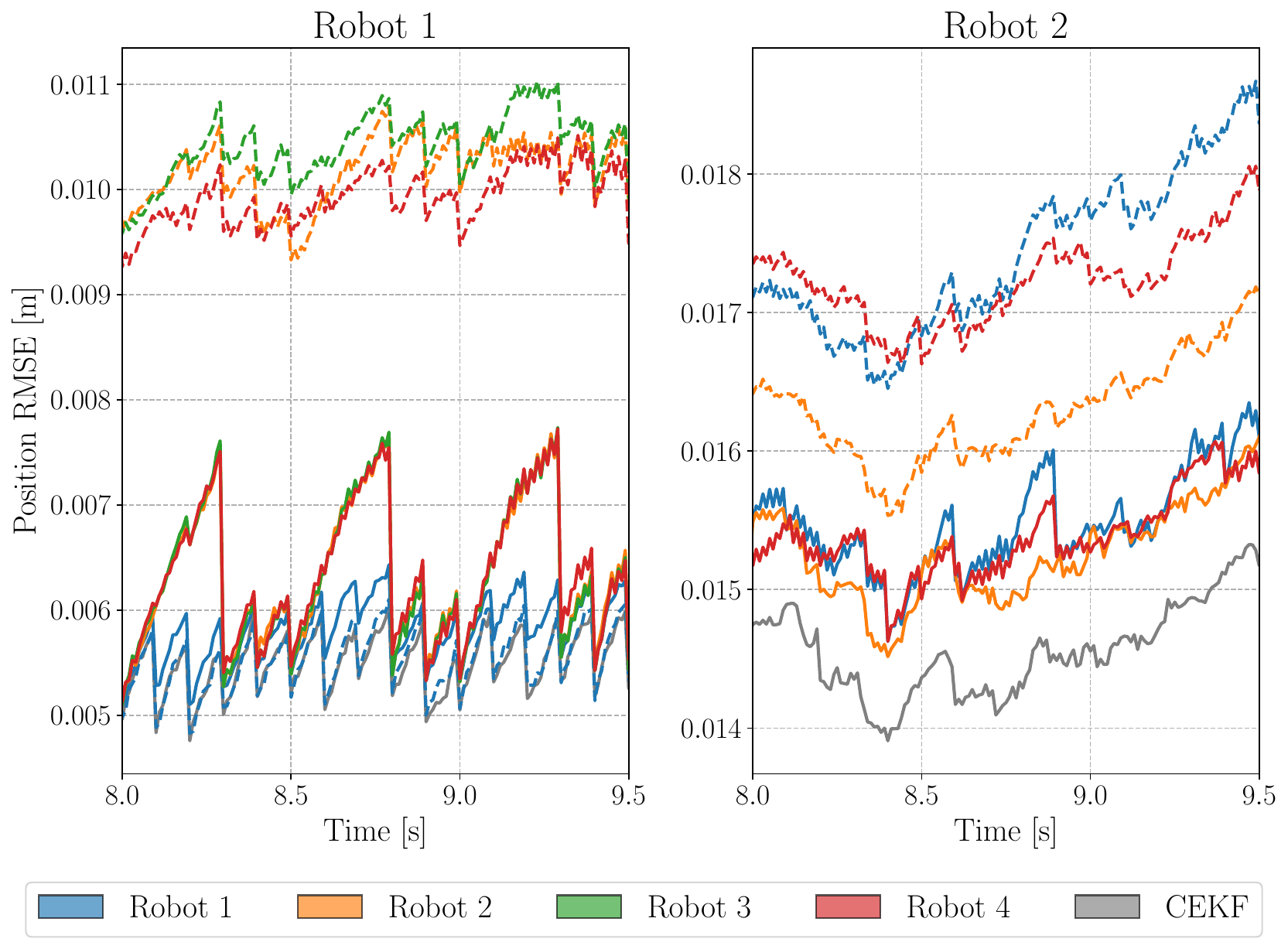}
    \caption{The S2 position RMSEs of Robot 1 and Robot 2 as estimated by the proposed (solid) and SoTA (dashed) distributed estimators, over $M=200$ trials, zoomed into a $1.5~\si{s}$ window to differentiate the methods.}
    \label{fig:RMSE_comparison_S2}
    \vspace{-3pt}
\end{figure}
Fig.~\ref{fig:RMSE_comparison_S2} shows the position RMSEs of Robot 1 and Robot 2 from all estimators for S2 over a $1.5~\si{s}$ window. For both robot estimates, it is clear that the proposed method produces estimates that are more in consensus than the SoTA, as the RMSEs overlap more. It is also clear that the proposed method produces more accurate estimates than the SoTA method, as they are closer to the centralized solution. The exception to this is Robot 1's estimate of itself, where the SoTA method performs marginally better than the proposed, at the expense of a poorer estimate of Robot 2's position. This is due to Robot 1 being capable of observing its global pose as it is equipped with GPS, and $\omega = 0.99$ giving little weight to the neighbouring estimates during consensus updates for the SoTA method. This trend occurs throughout the entire trajectory.

\subsection{Experimental Results}
\begin{figure*}[!t]
    \centering
    \begin{subfigure}{0.496\textwidth}
        \centering
        \includegraphics[width=\linewidth]{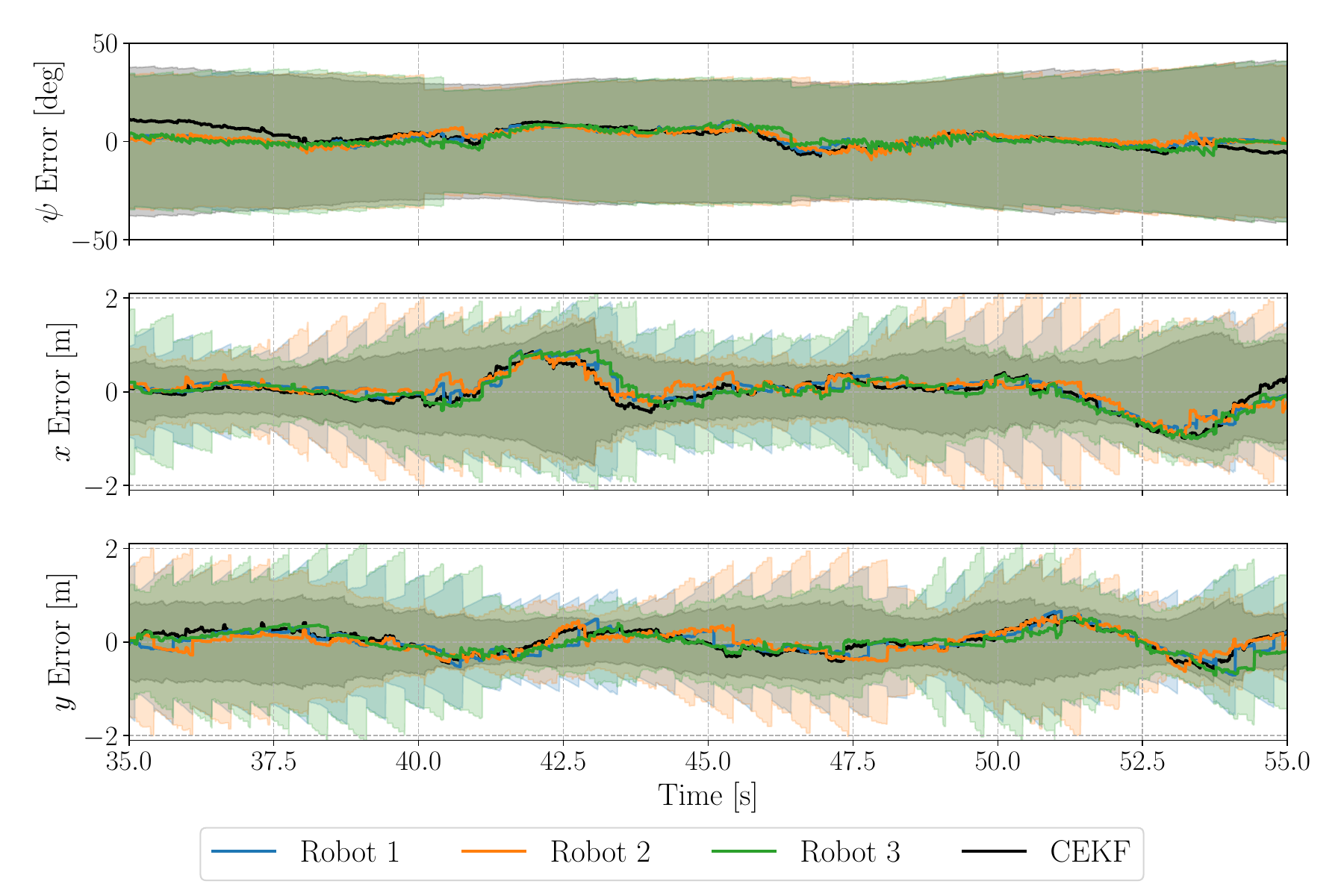}
        \caption{Proposed w/o Passive}
        \label{fig:error_prop_wo_psv}
    \end{subfigure}
    \hfill
    \begin{subfigure}{0.496\textwidth}
        \centering
        \includegraphics[width=\linewidth]{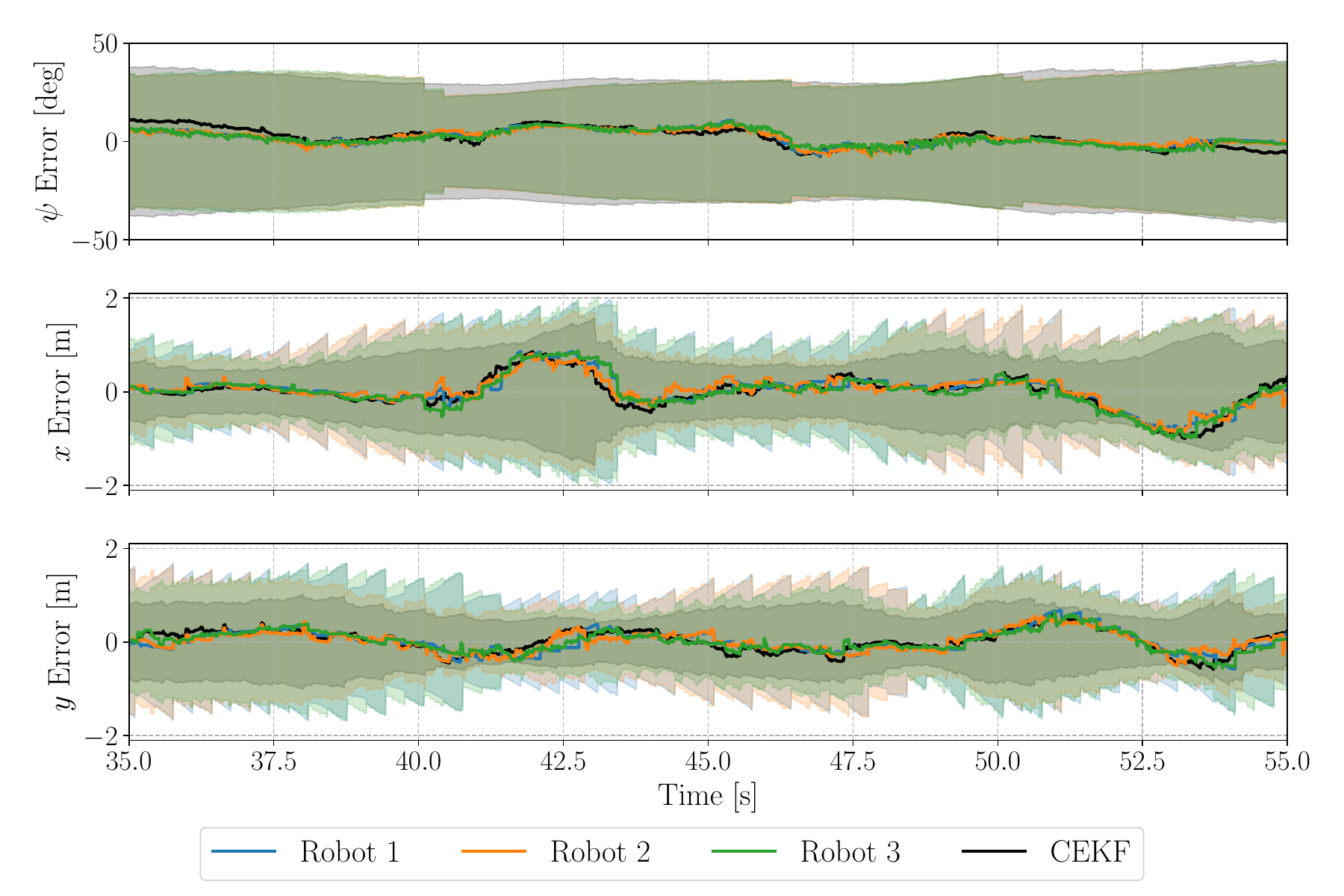}
        \caption{Proposed w/ Passive}
        \label{fig:error_prop_w_psv}
    \end{subfigure}
    \caption{Error plots and $\pm3\sigma$ bounds of Robot 1 by all estimators for experiment \texttt{default\_3\_random\_0b}, comparing the proposed method with and without passive listening over a $20~\si{s}$ window, for a consensus frequency of $3~\si{Hz}$.} 
    \label{fig:exp_error_plots}
    \vspace{-3pt}
\end{figure*}

\begin{figure}[h]
    \centering
    \includegraphics[width=\linewidth]{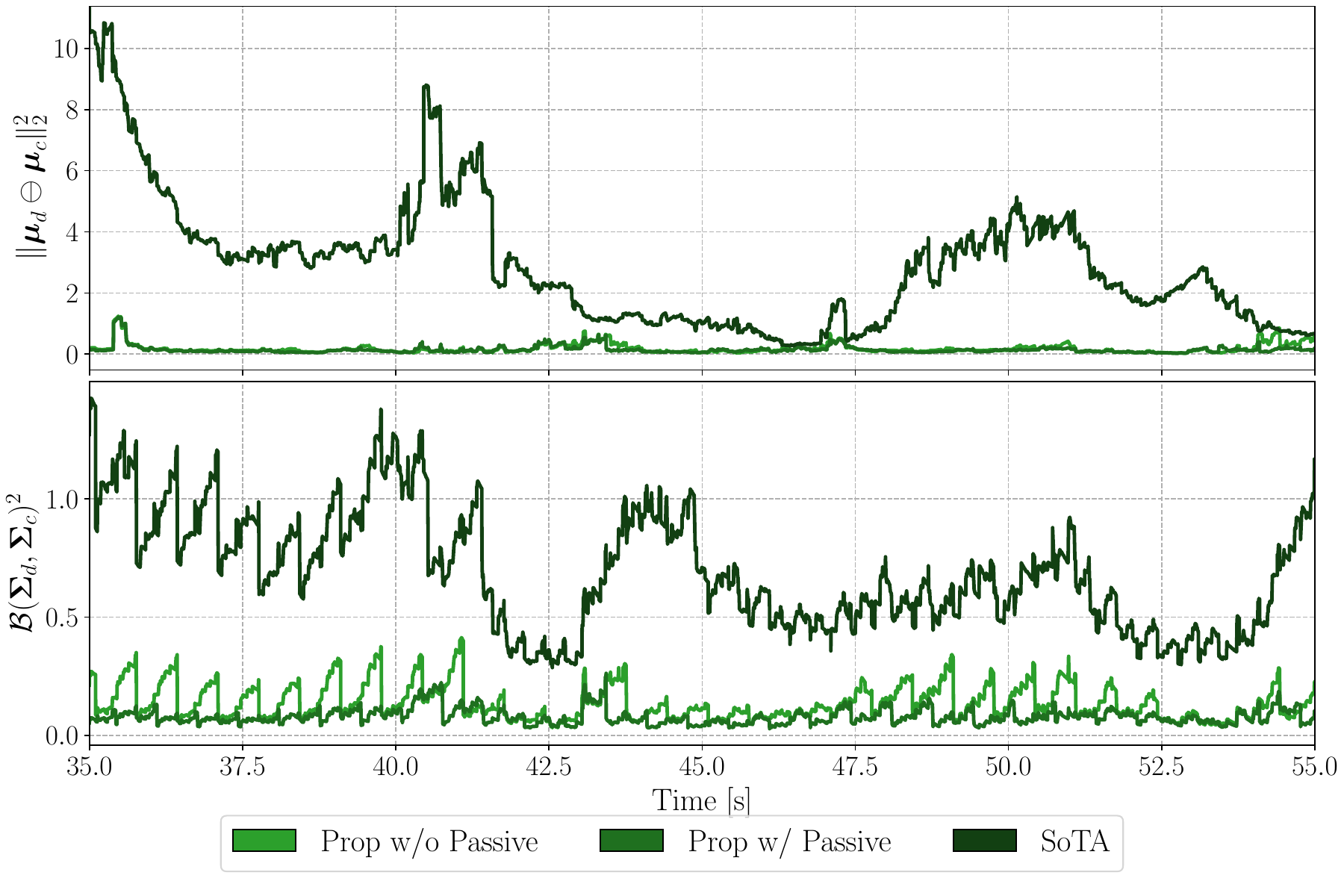}
    \caption{A $20~\si{s}$ window comparing the component-wise 2-Wasserstein metrics of Robot 3's estimates between the proposed and SoTA methods for the \texttt{default\_3\_random\_0b} experiment, for a communication frequency of $3~\si{Hz}$.}
    \label{fig:wass_exp}
    \vspace{-3pt}
\end{figure}

The MILUV dataset~\cite{MILUV_IJRR_2026} is used to experimentally validate the proposed method. The dataset consists of a team of three quadrotors flying in an indoor environment, each equipped with an IMU, an altimeter, and two UWB transceivers per robot, enabling robot-to-anchor and inter-robot ranging. In this scenario, the same inter-robot range measurement is available to both robots. Robots 2 and 3 are set to range to six anchors of known positions, while Robot 1 only gets global positioning information through shared state exchange. The edges of the communication graph are $\mc{E}=\{(1,2), (1,3)\}$. The robots perform inter-robot ranging and communication based on the communication graph. The local state of each robot $i$ is $\mc{X}_{i;i} = (\mbf{T}_{ai}, \mbf{b}_i) \in SE_2(3) \cross \mathbb{R}^6$, where $\mbf{T}_{ai}$ is the extended pose and $\mbf{b}_i$ is the gyroscope and accelerometer biases. The full state includes the neighbour states, $\mc{X}_{j;i}$ $\forall j \in \mc{N}_i$.
Fig.~\ref{fig:exp_error_plots} shows a $20~\si{s}$ window of the $\pm 3\sigma$ error plots of all robots' estimate of Robot 1's yaw and planar position, for the proposed method with and without passive listening at a consensus frequency of $3~\si{Hz}$. For both methods, the estimates are accurate as the errors are enveloped by the $\pm3\sigma$ bounds. However, it is clear that the proposed method with passive listening produces more consistent estimates, particularly for the position, than without, as the $\pm 3\sigma$ bounds are tighter and grow less frequently. This is attributed to the passive consensus updates, as without it, neighbouring estimators can only deadreckon their belief of Robot 1's state using received RMIs until their next scheduled consensus update. The benefit of passive listening is evident at lower communication frequencies, as there is a longer delay between consensus updates. For this particular sequence, it was observed that with a communication frequency of $2~\si{Hz}$, all distributed estimators diverged without passive listening, whereas with passive listening, the distributed estimates remained accurate and consistent.
Additionally, Fig.~\ref{fig:wass_exp} shows Robot 3's component-wise 2-Wasserstein metric for the proposed and SoTA methods over a $20~\si{s}$ window. For both the mean and covariance metrics, Robot 3's estimate from the proposed method is closer to the centralized solution. Additionally, the with passive case shows an improvement compared to without, as there is less time deadreckoning neighbour states without a consensus update. This shows that the proposed method produces more accurate and consistent estimates that are in consensus, and are closer to the centralized solution than the SoTA method. This is similarly seen for the other robots' estimates, which are not shown.
\subsection{Comparison to State-of-the-Art}
The main difference between the proposed and SoTA~\cite{cossette_decentralize_2024} methods is that the proposed only shares relevant shared states, whereas the latter shares the \emph{full} state during a consensus update. The SoTA method performs a CI step on the \emph{full} state, inflating unique states that are not a function of the consensus pseudomeasurement, producing more conservative estimates. The proposed method is thus lighter weight in communication, more scalable to larger teams and produces more accurate and consistent estimates, while not needing to implement user-defined pseudomeasurements.
    
    \section{Conclusion}\label{sec:conclusion}
In this paper, a novel communication-efficient consensus-based distributed filtering framework for collaborative localization on MLGs was developed, utilizing both preintegrated odometry sharing and state sharing through inter-robot communication. Additionally, a novel passive listening-based consensus update was also introduced for non-communicating neighbours to update their local estimates based on overhearing shared state exchanges, showing improvements to distributed estimates from more frequent consensus updates. The proposed method was validated in both simulation and experimental settings, showing near centralized performance in accuracy and particularly in consistency compared to the SoTA approach~\cite{cossette_decentralize_2024}. Future work includes extending the framework to an optimization-based approach, and handling dynamic communication graphs.  
    \section*{Acknowledgments}
Microsoft Copilot was used in a limited capacity to provide technical suggestions and draft review. All suggestions were carefully vetted. The authors take full responsibility for the paper's content.  
    \printbibliography
\end{document}